# AI Approach for MRI-only Full-Spine Vertebral Segmentation and 3D Reconstruction in Paediatric Scoliosis

Nathasha Naranpanawa[1,2], Maree T. Izatt[1,3], Robert D. Labrom[1,3], Geoffrey N. Askin[1,3], J. Paige Little[1,2,3]

[1]Biomechanics and Spine Research Group, School of Mechanical, Medical, and Process Engineering, Faculty of Engineering, Queensland University of Technology, Brisbane

[2]Centre for Biomedical Technologies, School of Mechanical, Medical and Process Engineering, Faculty of Engineering, Queensland University of Technology, Brisbane

[3]Orthopaedics Department, Queensland Children's Hospital, Brisbane, Australia

**Abstract**

Magnetic resonance imaging (MRI) is favoured over computed tomography (CT) for paediatric imaging generally due to the absence of ionising radiation; however, its clinical use in paediatric spine deformity remains largely limited to screening for and diagnosing neuroaxial abnormalities. Detailed three-dimensional (3D) anatomical assessment, complex congenital deformity evaluation, and surgical navigation still rely on CT, as MRI-based 3D spine reconstruction is currently manual, time-consuming, and constrained by the lack of high-resolution, labelled full-spine 3D-MRI datasets except for those specifically requested in the research setting. Consequently, accurate automated thoracolumbar spine 3D-MRI reconstruction has remained unattainable. This study presents an artificial intelligence (AI) framework that enables accurate, automated thoracolumbar spine vertebral segmentation and 3D reconstruction using MRI alone by leveraging historical adolescent scoliosis CT datasets through pseudo-MRI synthesis.

To address the lack of full-spine 3D-MRI datasets, historical low dose thoracolumbar spine CT scans of patients with adolescent idiopathic scoliosis (AIS) were transformed into 'MRI-like' images using a generative adversarial network (GAN) while preserving AIS-specific characteristics. These synthetic thoracolumbar spine MRIs were combined with an existing, manually labelled thoracic 3D-MRI research dataset to train a U-Net–based segmentation algorithm for the automated vertebral segmentation of the entire thoracolumbar spine from 3D-MRI.

The enhanced AI algorithm accurately segmented all thoracolumbar vertebral levels (T1-L5) and produced continuous 3D thoracolumbar spine reconstructions suitable for interrogation of the bony deformity. Compared with a baseline algorithm trained only on the available research dataset of thoracic only 3D-MRI, the enhanced AI algorithm achieved improved Dice accuracy (86% vs 88%), and drastically reduced processing time (~1 hour vs <1 minute). Crucially, the generative AI algorithm preserved AIS-specific anatomical features – including curvature, vertebral rotation, and sagittal alignment parameters– enabling reliable thoracolumbar spine 3D representation from a 3D-MRI scan.

This work demonstrates, for the first time, a scalable AI approach that enables accurate, fully automated thoracolumbar spine 3D reconstruction from MRI alone in AIS patients. By bringing MRI-based 3D bony assessment closer to the capabilities previously achievable only with CT, this method supports radiation-free deformity evaluation, pre-operative planning, and surgical navigation which is highly desirable for paediatric spine care.

**Introduction**

Magnetic resonance imaging (MRI) is favoured over computed tomography (CT) for paediatric imaging generally because it enables detailed visualisation of vertebrae and soft tissues without any exposure to ionising radiation (1–5). This advantage is particularly important in paediatric spine deformity, where patients are required to undergo repeated imaging (predominantly X-Ray/EOS) over many years to monitor the deformity during growth and brace treatment, for surgical planning, intra-operative monitoring and postoperative follow-ups (6). Compared with CT and radiographic imaging, MRI has the potential to eliminate much of the cumulative radiation dose while additionally providing excellent depiction of the spinal cord and nerve roots, intervertebral discs, and surrounding soft tissues, making it a safer modality for long-term spine care in children and adolescents (7–9).

Despite these benefits, routinely acquired clinical MRI sequences at most centres are primarily ordered as part of surgical planning and are optimised for detecting neural axis abnormalities (10,11). These clinical MRI sequences lack the spatial resolution required for high-fidelity three-dimensional (3D) reconstruction of osseous anatomy. While sufficient for qualitative assessment, standard clinical MRI acquisitions have coarse slice spacing and limited resolution, precluding accurate 3D vertebral reconstruction and detailed assessment of bony morphology.

3D spine models are increasingly relied upon by surgeons and clinicians to define deformity details, plan surgical corrective procedures and for surgical navigation (12,13). However, a major barrier to generating such precise bony models from MRI using artificial intelligence (AI) is the absence of labelled, high-resolution, full-spine 3D-MRI datasets of paediatric populations for training (14,15). Although high-resolution 3D-MRI datasets of adult spines exist (16–19), AI algorithms trained on these data do not generalise well to accurately segment spines of patients with paediatric scoliosis and spine deformities, owing to substantial differences in spinal length, growth-related variation, and deformity-specific vertebral morphology (20,21). Creating suitable 3D spine models would require high resolution 3D-MRI with finer slice spacing, and manual, slice-by-slice annotation of every vertebra in all three dimensions, a highly specialised and time-intensive task. As a result, large-scale labelled, full spine 3D-MRI datasets suitable for training AI algorithms for paediatric scoliosis are essentially unavailable currently.

In contrast, historical full-spine CT datasets of paediatric deformity patients, acquired in the past to assist with surgical planning, provide detailed vertebral segmentations and rich anatomical information of AIS patients' spines. However, CT and MRI differ fundamentally in image appearance, contrast mechanisms, and noise patterns (Figure 1). AI algorithm trained on CT data do not generalise well to detecting bony margins on MRI, a limitation known as cross-modality domain shift (22,23). This modality mismatch is particularly problematic in scoliosis where accurate representation of deformity-specific bony anatomy is critical. As a result, existing CT-labelled datasets cannot be directly leveraged for AI-generated MRI-based bony segmentation, especially without paired CT-MRI data.

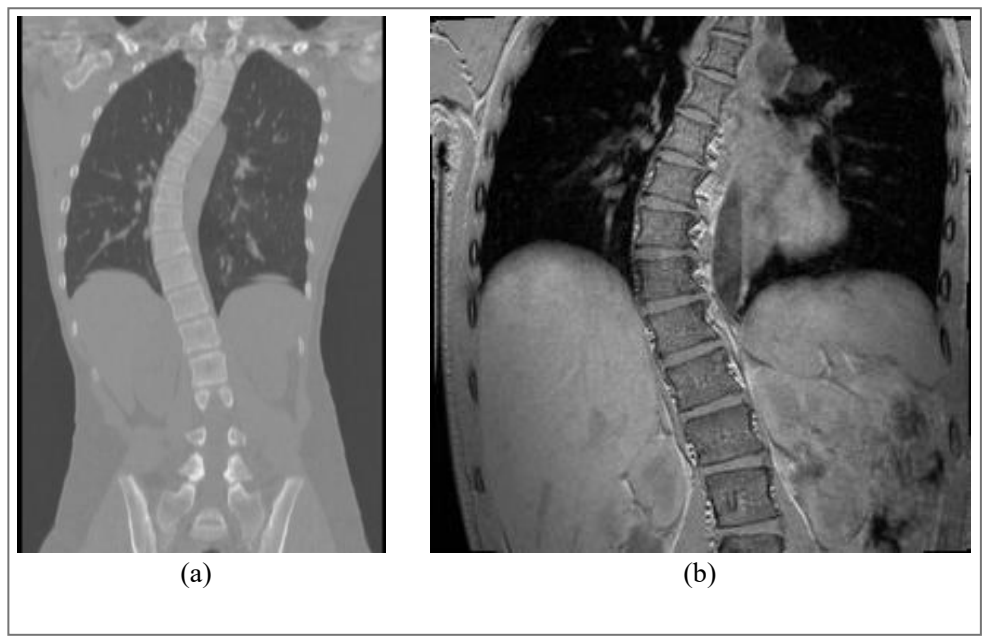

*Figure 1: Representative central two-dimensional (2D) slices extracted from full 3D volumes of two adolescent idiopathic scoliosis (AIS) patients' scans: (a) thoracolumbar computed tomography (CT) and (b) thoracic magnetic resonance imaging (MRI). The images highlight differences in contrast, resolution, and visualisation of osseous anatomy between modalities.*

To advance MRI toward a CT-equivalent tool for 3D spine assessment, our research group previously developed an AI method capable of reconstructing the thoracic spine from thoracic 3D-MRI in both healthy adolescents and patients with AIS, with MRI data acquired from an earlier research project (14,20,21,24). While this work demonstrated the feasibility of MRI-based 3D vertebral reconstruction, extension of the approach to the lumbar spine was not feasible due to the absence of labelled lumbar vertebrae in our existing thoracic 3D-MRI dataset. In addition, segmentation at inference required extensive post-processing, resulting in runtimes of up to an hour per case, which limited its practicality for routine clinical use. Consequently, despite establishing an important proof-of-concept, accurate, efficient, and fully automated full-spine 3D reconstruction from MRI remained unattainable.

This limitation directly motivated the present work, which seeks to overcome the lack of full thoracolumbar 3D-MRI data and enable scalable, rapid, and anatomically consistent 3D spine reconstruction from MRI. This study addresses this challenge by introducing an AI framework that bridges the gap between CT and MRI through synthetic image generation, without requiring paired CT and MRI data from each patient. Specifically, MRI-like images are generated from accurately labelled CT data, producing "pseudo-MRI" volumes that retain precise vertebral morphology from thoracolumbar scoliosis CT while exhibiting the MRI-like appearance required for training. This approach enables reuse of detailed CT-derived vertebral labels for training MRI-based AI algorithms, despite the absence of true high-resolution thoracolumbar 3D-MRI data.

The overarching goal is to enable accurate, automated full-spine bony segmentation using 3D-MRI alone at inference, while leveraging historical CT datasets exclusively during algorithm training. To our knowledge, this is the first study to address AIS vertebral segmentation by combining pseudo-MRI synthesis from historical CT with MRI-based segmentation training, enabling anatomically coherent 3D reconstruction in a setting where labelled full-spine MRI data do not exist. By doing so, this work aims to support radiation-free, high-quality full-spine 3D assessment for paediatric patients with AIS, with the potential to replace CT for deformity evaluation, surgical planning, and intra-operative navigation in future clinical practice.

**Methods**

This work utilised two existing imaging datasets: (i) a set of historical low-dose T1–L5 thoracolumbar CT scans, and (ii) a T1-weighted thoracic-only 3D-MRI dataset acquired for an earlier scoliosis progression research study. The historical dataset (Ethics MHS 25276) was comprised of 3D CT volumes from 91 AIS patients (82 females and 5 males, mean age 15.4±3.9 years). Vertebral annotations for these CT scans were a combination of manual labels and automatically generated labels through the TotalSegmentator software (version 2.5) (25). The thoracic 3D-MRI dataset (Ethics CHQHHS 24390) was comprised of 60 thoracic spine 3D-MRI scans from 25 AIS patients (mean age 12.4±1.3 years). The vertebral annotations for the 3D-MRI volumes were manually generated by trained personnel using Amira software (version 6.7).

To address the absence of comprehensive labelled lumbar region 3D-MRI datasets, the historical low-dose T1–L5 thoracolumbar CT dataset with existing vertebral annotations were leveraged. Although CT is currently used very selectively in paediatric deformity due to radiation exposure, this archival dataset provided detailed vertebral labels spanning both the thoracic and lumbar spine, granting access to a wider field of view of the vertebral continuity and patterns in deformed and scoliotic paediatric spines.

Figure 2 illustrates the workflow for training of the AI models with these datasets, and inference using the trained models for 3D reconstruction.

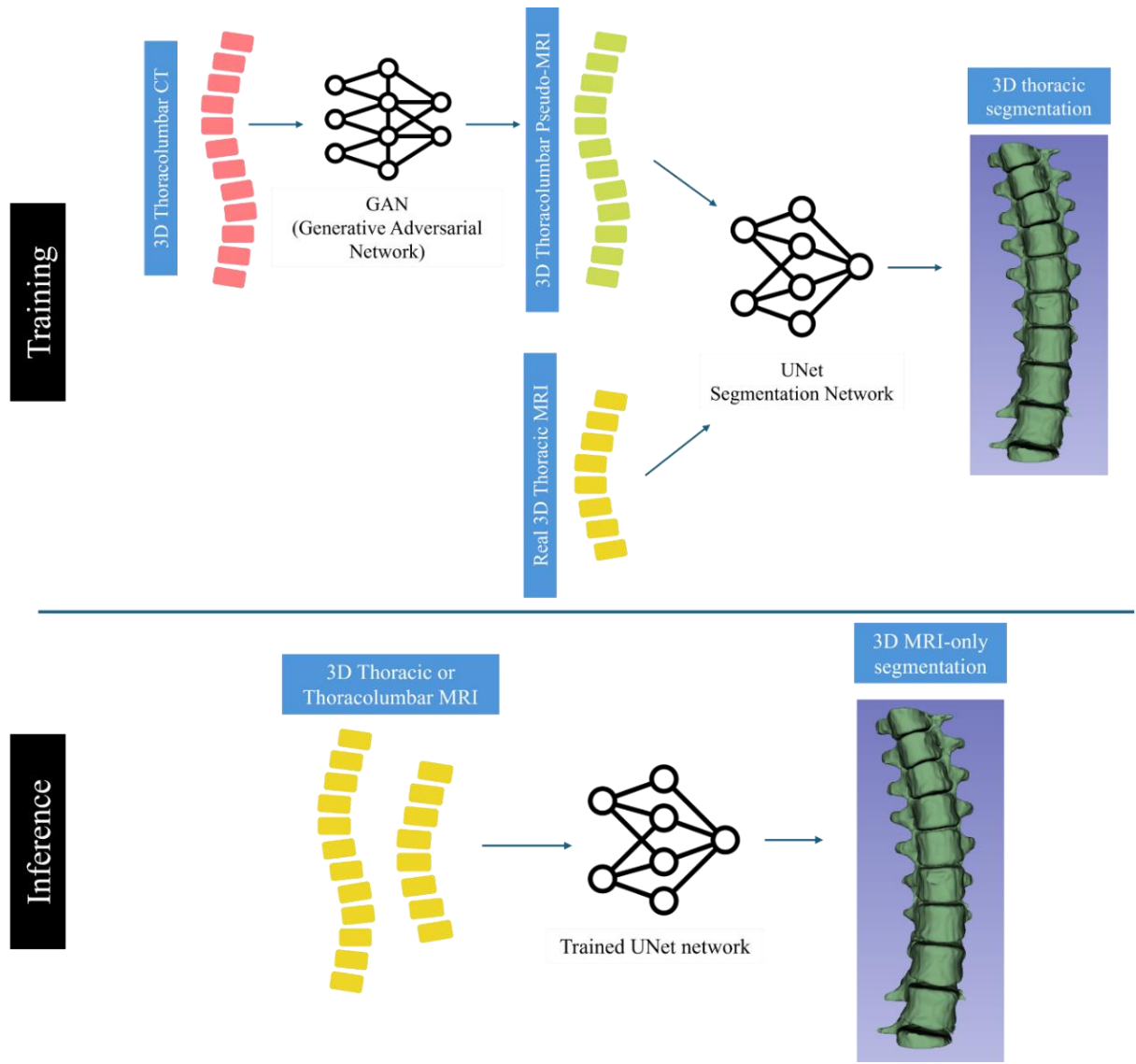

*Figure 2: Overview of the proposed AI framework for MRI-based spine segmentation. Upper panel: During training, historical 3D thoracolumbar CT volumes are converted to pseudo-MRI using a generative adversarial network (GAN) and combined with real thoracic 3D-MRI to train a U-Net segmentation network. Lower panel: At inference, the trained U-Net model generates vertebral segmentations and 3D reconstructions using thoracic or thoracolumbar MRI as the sole input, with no CT required.*

(a) Pseudo-MRI generation with generative adversarial network

Firstly, a generative adversarial network (GAN) (26) was trained to transform the thoracolumbar 3D CT volumes into synthetic MRI-like images that closely replicate the visual characteristics of MRI. Training was informed by the unpaired thoracic-only 3D-MRI dataset. Although limited to the thoracic region and occasionally L1 vertebra, these research derived 3D-MRIs provided authentic MRI contrast and patient-specific deformity patterns critical for anatomical realism. This enabled the GAN to learn MRI-like soft-tissue contrast while preserving the precise bony morphology and deformity-specific anatomy inherent to the original CT scans. This process effectively bridged the appearance gap between CT and MRI, retaining the detailed anatomical richness of CT while generating images compatible with

MRI-based AI algorithms, resulting in a synthetic full thoracolumbar pseudo-MRI dataset that can be utilised for training.

(b) MRI segmentation and reconstruction

The synthetic thoracolumbar pseudo-MRI output from the GAN were subsequently combined with the thoracic 3D-MRI dataset. Together, the combined datasets offered both complete bony anatomical coverage of the thoracolumbar spine, derived from the pseudo-MRI volumes, and true MRI signal characteristics from the thoracic 3D-MRI acquisitions.

A U-Net–based deep learning architecture (27) was then trained on the combined dataset to perform automated bony segmentation across the entire thoracolumbar AIS spine. During training, the segmentation model learned to integrate information from both the synthetic thoracolumbar pseudo-MRI and the true thoracic 3D-MRI domains, enabling it to infer and reconstruct lumbar vertebral anatomy even when the input MRI data contained thoracic coverage only. The network was optimised to maintain continuity between vertebral levels and to accurately capture scoliosis curve morphology, vertebral rotation, and bony boundaries relevant for clinical assessment and surgical application.

Model performance was evaluated on a held-out subset of the AIS 3D-MRI scans (5 volumes) not used during training. Evaluation focused on the anatomical consistency of thoracic spine segmentation and the extent to which inclusion of pseudo-MRI data improved segmentation accuracy and generalisability in the thoracic region. Quantitative assessment was performed using the Dice similarity coefficient as the primary metric.

(c) MRI-only inference

At inference, the trained U-Net model operates exclusively on MRI data and does not require CT input. Given a thoracic or thoracolumbar 3D-MRI volume, the network directly generates vertebral segmentation masks, which are then assembled into a continuous 3D reconstruction of the spine. Although the available inference data were limited to thoracic MRI coverage, the segmentation model leveraged the anatomical priors learned from the pseudo-MRI–enhanced training to produce anatomically consistent vertebral segmentation across the entire imaged region (T1-L5). Importantly, inference is fully automated and computationally efficient, enabling rapid generation of segmentation outputs without any post-processing, thereby supporting practical integration into clinical workflows.

**Results**

A baseline U-Net model trained solely on manually labelled thoracic 3D-MRI achieved a Dice similarity coefficient of 86.1% for thoracic vertebral segmentation. While effective within the limited thoracic field of view, this segmentation algorithm demonstrated restricted generalisability to adjacent spinal regions due to the absence of cervical and lumbar region training data.

The GAN-based pseudo-MRI synthesis successfully preserved patient-specific spinal deformities while generating realistic MRI-like appearance across the entire thoracolumbar spine (Figure 3), without the requirement of any paired thoracolumbar MRI information. Visual inspection confirmed that vertebral boundaries, curvature patterns, and relative intervertebral relationships characteristic of AIS were maintained following modality translation.

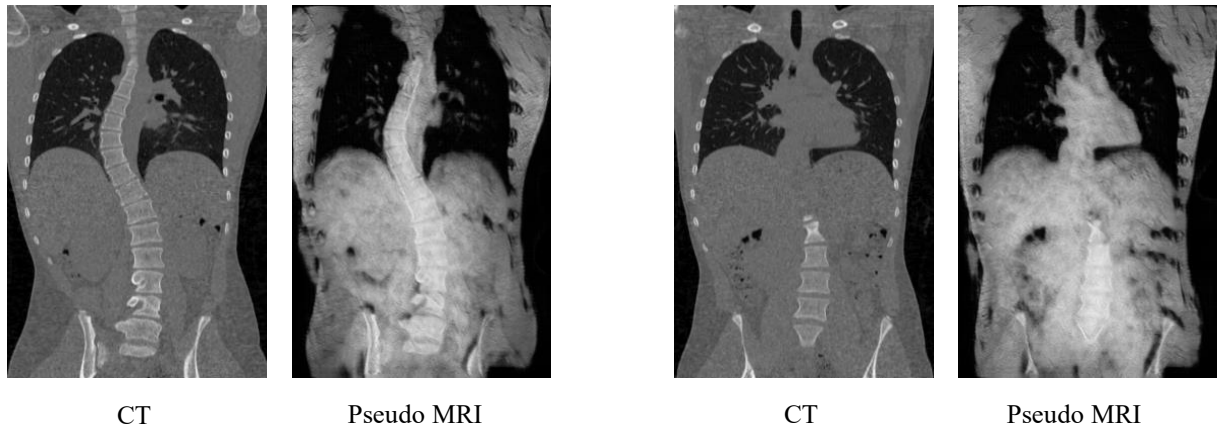

CT            Pseudo MRI            CT            Pseudo MRI

*Figure 3: Example coronal CT images of AIS spines and their corresponding pseudo-MRI translations, demonstrating effective cross-modality synthesis that retains deformity-specific bony anatomy while approximating MRI appearance.*

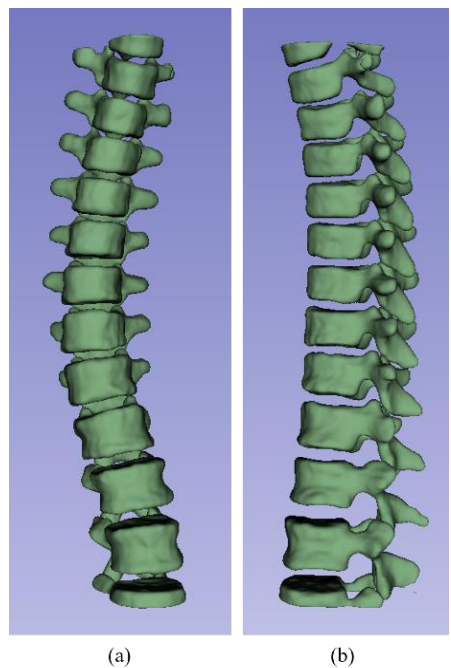

(a)            (b)

*Figure 4: An example of the (a) coronal, and (b) sagittal views of the thoracic region of an AIS spine segmented and reconstructed in 3D using the enhanced AI segmentation model is shown. As full-spine 3D-MRI is not routinely acquired in the clinical setting, we currently demonstrate thoracic-only AIS spine reconstructions from our thoracic 3D-MRI research dataset, although the AI algorithm is designed for thoracolumbar AIS spine inference.*

Quantitatively, incorporation of pseudo-MRI data enabled the enhanced spine segmentation model to achieve an improved Dice score of 87.5% for thoracic segmentation. Significantly, at inference, both baseline and enhanced segmentation models were able to provide accurate 3D reconstructions (Figure 4) in less than a minute of end-to-end processing time, with no requirement for further post processing.

Comparative analysis demonstrated that AI segmentation algorithm trained with pseudo-MRI data achieved higher AIS thoracic spine segmentation accuracy and produced smoother, more coherent 3D bony reconstructions than the segmentation algorithm trained on thoracic 3D-MRI data alone. Importantly, the final trained AI model is capable of generating a full thoracolumbar or thoracic spine 3D bony reconstruction for AIS patients using MRI as the sole input at inference (Figure 2), supporting its intended use as a radiation-free, clinically deployable tool for AIS surgical planning and intra-operative navigation.

**Discussion**

This study demonstrates that pseudo-MRI synthesis can effectively bridge the modality gap between CT and MRI, enabling automatic and fast 3D vertebral segmentation of the full thoracolumbar region of paediatric scoliosis spines using only 3D-MRI at inference. By leveraging historical CT datasets with reliable vertebral annotations, the proposed framework addresses one of the most significant challenges in MRI-based spine artificial intelligence: the scarcity of labelled, high-resolution full thoracolumbar 3D-MRI data. Importantly, the AI algorithm is now sufficiently trained such that CT data are no longer required at inference, supporting a radiation-free workflow that aligns with current paediatric imaging priorities, even in the absence of true full-spine 3D-MRI acquisitions for its training.

Experiments revealed two key findings. First, incorporation of pseudo-MRI data resulted in measurable improvements in thoracic vertebral segmentation performance compared with the earlier MRI-only baseline, with higher Dice similarity coefficients and improved boundary accuracy. Crucially, these gains were accompanied by improved segmentation continuity and reduced inter-vertebral discontinuities, which are essential for producing coherent 3D vertebral reconstructions suitable for quantitative analysis rather than isolated per-slice segmentations.

Second, and more substantively, pseudo-MRI-enhanced training enabled anatomical generalisation beyond the limited thoracic field of view available in the real MRI training dataset. AI models trained exclusively on thoracic 3D-MRI were unable to produce reliable segmentations outside the thoracic region, whereas the inclusion of pseudo-MRI data enabled the network to infer anatomically consistent vertebral structures extending into adjacent spinal regions when applied to true thoracolumbar MRI. While this does not yet constitute direct segmentation of the thoracolumbar spine – which is not currently available – it demonstrates the segmentation model's capacity to internalise global spinal anatomy and deformity patterns rather than overfitting to a restricted field of view.

The pseudo-MRI synthesis preserved AIS-specific deformities, including curvature magnitude, vertebral alignment and wedging, and rotational patterns, which is essential for clinical relevance. Unlike anatomically rigid regions where synthetic image generation has been more extensively studied, the spine represents a deformable structure with substantial inter-subject variability. The ability of the generative adversarial network (GAN) to maintain deformity-specific anatomy and detail suggests that the learned cross-modal translation captures structurally meaningful features rather than superficial intensity mappings.

Compared to the previous work within the research group (14), current work drastically reduces end-to-end processing time from approximately one hour to under one minute, highlighting its potential for near–real-time clinical use. The improved AI framework automatically generates a high-quality 3D bony reconstruction and requires minimal manual input or post-processing. This improvement in both accuracy and efficiency supports integration into routine clinical workflows, where rapid generation of consistent 3D spine reconstructions is essential.

Importantly, the framework developed in this work is inherently forward-compatible with future advances in patient imaging. If high-resolution thoracolumbar or full-spine 3D-MRI becomes routinely acquired in AIS care – whether through faster imaging sequences, improved patient tolerance, or evolving clinical guidelines – the enhanced algorithm will be well positioned to directly incorporate these richer inputs. In such a scenario, the AI system could transition from inferring thoracolumbar anatomy to directly segmenting it from 3D-MRI, with the potential for further gains in accuracy and clinical utility for deformity assessment and surgical navigation.

A key contribution of this work is that it enables MRI-only 3D AIS spine reconstruction, without reliance on proprietary commercial software or CT-based workflows. At present, clinically available 3D spine reconstruction tools are largely CT-dependent and embedded within commercial planning or navigation systems, limiting accessibility and necessitating radiation exposure in paediatric patients (28–31). By contrast, the proposed framework provides an open, data-driven pathway for generating anatomically coherent 3D spine models directly from MRI, using publicly available deep learning architectures trained on historical research-based datasets. This capability is particularly important in the paediatric spine deformity setting, where repeated imaging is common and necessary for their management throughout their growing years when deformity progression risk is highest. By removing the requirement for CT at inference and avoiding dependence on commercial reconstruction platforms, this work establishes a practical foundation for non-commercial, MRI-only 3D spine reconstruction that can be integrated into research, clinical decision-making, and future digital care pathways.

One important potential application of accurate MRI-derived 3D spine reconstructions is integration with spinal navigation and robotic systems, which currently rely on CT-based bony anatomy to guide pedicle screw placement (32,33). Achieving vertebral segmentation of sufficient accuracy from 3D-

MRI would represent a major advance toward reducing reliance on CT for intra-operative navigation. The use of MRI in surgical navigation systems and robotics would eliminate this additional and substantial radiation exposure in a population of paediatric patients who already have an existing cumulative radiation exposure as part of their standard care throughout their childhood.

The workflow is inherently scalable across metropolitan imaging services and, with appropriate infrastructure, selected regional centres. Tertiary spinal units could utilise the system to support preoperative planning and multidisciplinary case discussions, while outreach and telehealth services could leverage 3D reconstructions to facilitate remote consultation with specialist teams.

Beyond AIS, the approach has potential applicability to other paediatric spinal deformity conditions and adult degenerative spine deformity, where reducing radiation exposure and improving anatomical visualisation remain important clinical goals. However, some important limitations merit consideration. The available dataset comprised exclusively right-convex curves in patients with adolescent idiopathic scoliosis, and accordingly the current framework has been validated only for right-curved AIS spines. Extension of the algorithm to segment and reconstruct left-sided curves and other scoliosis subtypes will require additional representative training data. Furthermore, the AI algorithm was developed specifically for idiopathic scoliosis and has not been evaluated on other paediatric spinal conditions, such as congenital or neuromuscular scoliosis, nor on adult degenerative deformities, where anatomical characteristics, deformity patterns, and imaging appearances may differ substantially.

Future work will focus on prospective clinical evaluation and multi-site validation, as well as refinement of the AI framework toward whole-spine segmentation from C1 to S1 as imaging availability evolves. Integration with digital surgical planning and navigation platforms and advanced 3D visualisation tools will further support clinical adoption. Ultimately, this work aims to contribute toward a statewide, radiation-free, MRI-only digital pathway for automated and accurate paediatric scoliosis spine segmentation when the requisite imaging becomes more readily available.

**Conclusions**

This work introduces an artificial intelligence–based framework that enables automatic vertebral segmentation and 3D spine reconstruction in adolescent idiopathic scoliosis patients using 3D-MRI alone at inference. By addressing the combined challenges of cross-modality domain mismatch and the absence of labelled, high-resolution thoracolumbar MRI data, the proposed approach unlocks the value of existing CT resources to support radiation-free, MRI-based AI workflows.

Through quantitative evaluation we show that pseudo-MRI–enhanced training improves bony segmentation accuracy, enhances anatomical continuity, and enables generalisation beyond the restricted thoracic field of view we had available for real MRI AI training data. This represents a substantive advance in our development of a practical pathway toward clinically meaningful 3D spine reconstruction direct from 3D-MRI in paediatric scoliosis. While full-spine 3D-MRI are not routinely requested in the clinical setting, this work demonstrates that comprehensive and detailed 3D bony reconstruction from MRI is now technically achievable, a capability that did not previously exist.

More broadly, this study highlights the strategic role of synthetic modality generation as a means of extending the functional and anatomical scope of medical imaging AI, rather than serving solely as a tool for image augmentation. By enabling MRI-based 3D spine reconstruction without reliance on CT at inference, the proposed framework moves MRI closer to fulfilling roles traditionally reserved for CT which is highly desirable for patients of all ages. As MRI acquisition protocols and sequences continue to evolve, this approach offers a scalable foundation for safer, more comprehensive, and radiation-free spine care in paediatric populations, with potential applicability across a broad range of musculoskeletal imaging challenges.

## Acknowledgements

A portion of this work was funded by a grant from the AOA Research Foundation.

The authors sincerely thank the adolescents and their parents/carers for their generous participation in the research projects that contributed the data that made this work possible.